\documentclass[letterpaper, 10 pt, conference]{ieeeconf}  

\IEEEoverridecommandlockouts                              

\usepackage{graphicx} 
\usepackage{subfloat} 
\usepackage{epsfig} 
\usepackage{amsmath} 
\usepackage{amssymb}  

\DeclareMathOperator{\atantwo}{atan2}

\usepackage{tikz}
\usetikzlibrary{shapes,arrows,positioning,calc}
\newcommand\IEEEcopyrighttext{%
  \centering\footnotesize This work has been submitted to the IEEE for possible publication. Copyright may be transferred without notice, after which this version may no longer be accessible.}
\newcommand\IEEEcopyrightnotice{%
\begin{tikzpicture}[remember picture,overlay]
\node[anchor=south,yshift=-2.2cm] at (current page.north) {\fbox{\parbox{\dimexpr0.9\textwidth-\fboxsep-\fboxrule\relax}{\IEEEcopyrighttext}}};
\end{tikzpicture}%
}

\title{\LARGE \bf
Target Tracking via LiDAR-RADAR Fusion for Autonomous Racing
}

\author{Marcello Cellina, Matteo Corno and Sergio Matteo Savaresi
\thanks{All authors are with the Dipartimento di Elettronica, Informazione e Bioingegneria (DEIB),
        Politecnico di Milano, Via Ponzio 34/5, 20133 Milano, Italy
        {\tt\small \{marcello.cellina, matteo.corno, sergio.savaresi\}@polimi.it}}%
}

\begin{document}

\maketitle
\IEEEcopyrightnotice

\thispagestyle{empty}
\pagestyle{empty}

\begin{abstract}


High-speed multi-vehicle autonomous racing increases the safety and performance of road-going Autonomous Vehicles. Precise Vehicle Detection and Tracking from a moving platform is a key requirement for planning and executing complex autonomous overtaking maneuvers. To address this requirement, we have developed a latency-aware EKF-based Multi Target Tracking algorithm fusing LiDAR and RADAR measurements. The algorithm exploits the different sensor characteristics by explicitly integrating the range-rate in the EKF measurement function, as well as a priori knowledge of the racetrack during state prediction. It can handle Out-Of-Sequence Measurements via Reprocessing using a double state and measurement buffer, ensuring sensor delay compensation with no information loss. This algorithm has been implemented on Team PoliMOVE's autonomous racecar, and was validated experimentally by completing a number of fully autonomous overtaking maneuvers at speeds up to 275 km/h. 

\end{abstract}

\section{INTRODUCTION}

Multi-vehicle autonomous racing allows researchers to safely develop and test the software and hardware for Autonomous Vehicles (AVs) operating under a variety of at-the-limit conditions which would be dangerous to replicate with human drivers. The development of algorithms capable of reacting to high-speed unpredictable maneuvers of other vehicles will greatly enhance the safety of future commercial AVs. The Indy Autonomous Challenge (IAC), a multi-vehicle racing competition of self-driving cars operated by university researchers, is the optimal proving ground for the development of these technologies.

Vehicle Detection and Tracking (VDT) is a key enabling functionality of AVs, especially for the successful planning and execution of overtaking and defensive racing maneuvers. The key requirements for a VDT algorithm for this application lie in its accuracy and consistency in opponent pose and velocity estimation. The algorithm needs to run online, be aware of and compensate for sensor latencies in order to provide the most up-to-date possible estimate of the opponent state.

\begin{figure}[t]
        \centering
        \includegraphics[width=\columnwidth]{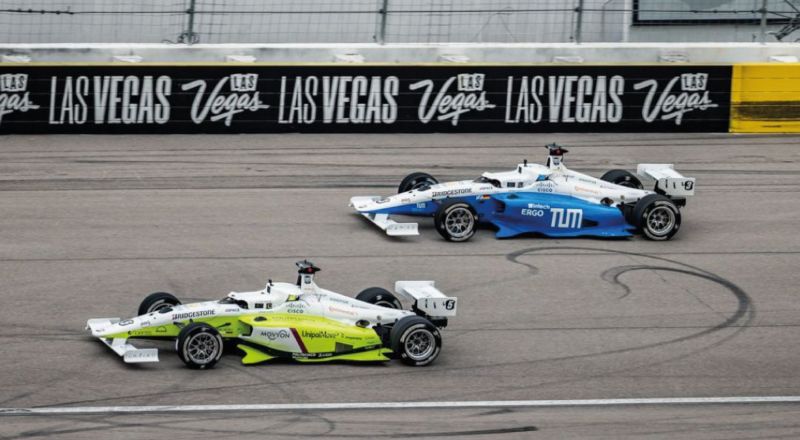}
        \caption{Team PoliMOVE's Dallara AV-21 (yellow) during an autonomous overtaking maneuver at the Las Vegas Motor Speedway, January 2023. Credits: \textit{Indy Autonomous Challenge}}
        \label{fig:intro}
\end{figure}

In this work, we present a latency-aware Multi-Target Tracking (MTT) algorithm, employing an Extended Kalman Filter (EKF) for state estimation with a reduced-state Constant Velocity and Turn Rate (CVTR) model and a double state and measurement buffer for Out-Of-Sequence Measurement (OOSM) management via Reprocessing. The MTT algorithm performs the fusion of pre-processed LiDAR and RADAR measurements coming from sensors mounted on the ego vehicle via explicit integration of the range-rate measurement into the EKF measurement equation together with a Yaw Rate Exogenous Input derived from the curvature of the track circuit.

The algorithm has been implemented and deployed on Team PoliMOVE's Dallara AV-21 and has proven its performance by completing a set of fully autonomous overtaking maneuvers at speeds up to 275 km/h.

The main contributions of this paper are:
\begin{itemize}
        \item A latency-aware architecture for an online EKF-Based MTT algorithm, allowing OOSM management without information loss.
        \item The inclusion of a priori track knowledge in the opponent motion prediction. 
        \item The explicit integration of the RADAR range-rate measurement in the EKF Measurement Function.
\end{itemize}

We experimentally validate the performance achieved by our contribution on both real data acquired during the 2023 IAC competition at the Las Vegas Motor Speedway, as well as on simulated data at the Autodromo Nazionale Monza. Ablation studies with respect to the RTK-GNSS ground truth demonstrate the contribution of each algorithmic module to the overall tracking performance. 

The remainder of the paper is structured as follows: in Section \ref*{sec:related_works}, we analyze the current state of the art and research gaps concerning LiDAR-RADAR Target Tracking for Autonomous Racing. In Section \ref*{sec:methodologies}, we describe in detail the structure and the components of the algorithms we developed, and then in Section \ref*{sec:experimental_results} we validate our claims through the analysis of experimental results. Finally, in Section \ref*{sec:conclusions}, we conclude our work discussing the achieved results and we investigate potential future developments.

\section{RELATED WORK}\label{sec:related_works}
Target tracking is the process of recursively estimating the dynamic states of one or more moving objects using sensor measurements. As a critical component of autonomous driving systems, it can be broadly classified into Multi-Target Tracking (MTT) and Extended Target Tracking (ETT) approaches, and then further divided depending on the sensors used and the vehicle dynamics estimator and model employed.

MTT algorithms maintain separate tracks, one for each object, assuming one measurement per object per timestep. Classical MTT frameworks, such as Multiple Hypothesis Tracking (MHT) \cite{dayangac2016target} and Joint Probabilistic Data Association (JPDA) \cite{fortmann1983sonar}, handle data association uncertainties. ETT methods address scenarios where a single extended object generates multiple measurements \cite{pieroni2024design}.

For state estimation, the Extended Kalman Filter (EKF) is a popular choice for its computational efficiency and handling of nonlinear measurements \cite{julier2004unscented}. More advanced model-based methods, such as the Unscented Kalman Filter (UKF) and Particle Filters (PF), have been developed for stronger nonlinearities or non-Gaussian distributions \cite{tian2022comparing}. Although computationally more intensive, these filters can offer significant accuracy improvements over the EKF in challenging scenarios.

The chosen motion model significantly impacts tracking performance for model-based estimators like EKF and UKF. Simple models like constant velocity (CV) and constant acceleration (CA) serve as basic assumptions for target dynamics \cite{li2003survey}. However, vehicles often perform complex maneuvers better represented by nonlinear models, such as Constant turn rate and velocity (CTRV) and constant turn rate and acceleration (CTRA) \cite{roth2014ekf}. Advanced modeling techniques, like the Interacting Multiple Model (IMM) estimator, combine multiple motion models to adapt dynamically to changing vehicle behavior \cite{mazor1998interacting}, although at the cost of increased computational and implementation effort.

Sensor fusion \cite{yeong2021sensor}, particularly combining LiDAR and RADAR, leverages complementary sensor strengths to improve tracking robustness and accuracy \cite{khaleghi2013multisensor}. LiDAR offers precise spatial positioning, while RADAR provides larger range and direct Doppler range-rate measurements. The authors in \cite{karle2023multi} demonstrate superior vehicle tracking performance by integrating LiDAR and RADAR data, although they do not explicitly integrate range-rate measurements.

Despite extensive research, current methods exhibit limitations in high-speed autonomous racing. Existing algorithms often neglect direct integration of RADAR Doppler range-rate measurements into state estimation, limiting velocity estimation accuracy. Additionally, most validations occur in urban or highway scenarios, lacking testing under the high dynamics and speeds of autonomous racing. These gaps underscore the necessity of specialized sensor fusion algorithms designed explicitly for high-speed racing environments.

\section{METHODOLOGY}\label{sec:methodologies}

In this section, we present our approach for target tracking via LiDAR-RADAR sensor fusion for autonomous racing. We developed a Multi-Target Tracking (MTT) algorithm using an Extended Kalman Filter (EKF) for track dynamics estimation enhanced with a measurement buffering system that compensates for sensor latency and manages Out-of-Sequence Measurements (OOSM). We model the vehicle dynamics using a Constant Velocity Turn Rate (CVTR) model and introduce a Yaw Rate Exogenous Input to further refine state estimation accuracy. A Finite State Machine (FSM) manages track initialization and termination, while data association is handled through a Global Nearest Neighbor (GNN) search.

We propose an algorithm that assumes availability of the following signals:

\begin{itemize}
        \item A smooth 2D absolute ego vehicle pose and velocity estimation \emph{e.g.}, from an INS/GNSS receiver
        \item A set of LiDAR-detected objects with 2D position measurements \emph{e.g.}, from a clustering algorithm \cite{cellina2025lidar}
        \item A set of RADAR-detected objects with 2D positions and range-rate measurements 
\end{itemize}

All the aforementioned signals come from sensors mounted on the moving sensing platform (ego vehicle), although the methods can be expanded to use signals coming from external sources like V2V communication. The sensors are supposed to be calibrated w.r.t. a common reference frame fixed on the ego vehicle.
We assume the sensor measurement delay to be known, even if time-varying. If no reliable delay measurement is available, an average constant value should be identified experimentally via suitable calibration procedures.

\begin{figure}
        \centering
        \includegraphics[width=\columnwidth]{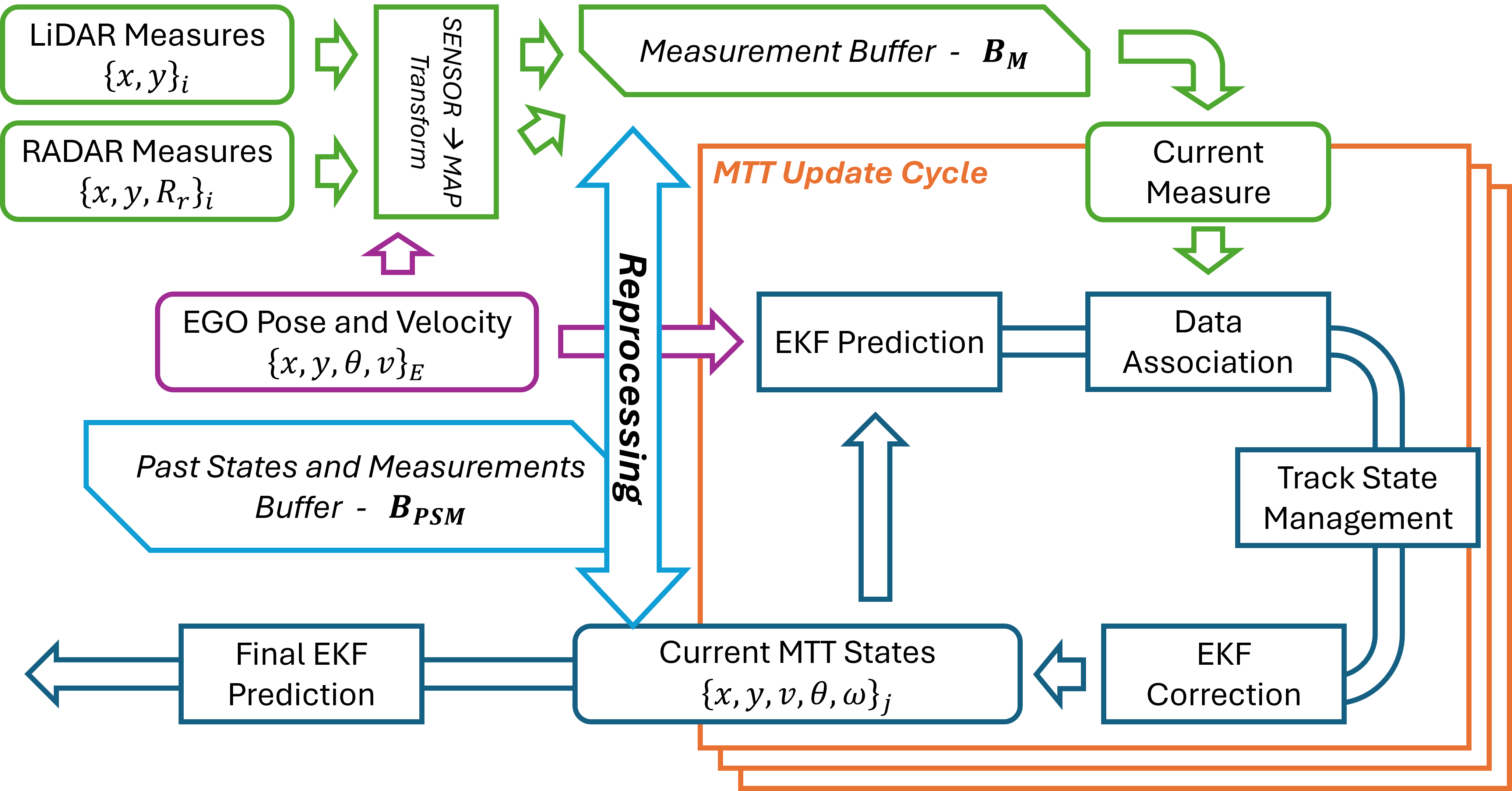}
        \caption{The proposed algorithm architecture with its main components}
        \label{fig:architecture}
\end{figure}

\subsection{Multi-Target Tracking with OOSM management}
In order to operate online, we implemented the algorithm as a fixed-rate ROS2 node with a \textit{Measurement Buffer} $B_M$. The main algorithm iterates at a fixed frequency of 33 Hz, higher than the scan frequency of both sensors. Incoming measurements between Target Tracking iterations are accumulated in $B_M$, which is implemented as an Ordered Map with the measurement timestamp $t_{meas}$ as the map key. 

At each algorithm iteration, we perform a full \textit{MTT Update Cycle}, composed of \textit{EKF Prediction, Data Association, Track State Management} and \textit{EKF Correction}, for every set of measurements in $B_M$, as shown in Figure \ref{fig:architecture}. Every measurement is removed from the buffer after its processing, and the next measurement is processed. The iteration ends when the buffer is empty. If the buffer was already empty at the beginning of the iteration, we still perform the Target Tracking cycle, even if only \textit{EKF Prediction} actually takes place, as the other stages aren't defined in the absence of new measurements. 

After processing a measurement, the MTT timestamp $t_{MTT}$ is updated to the timestamp of the most recent processed measurement, $t_{meas}^{old}$. If no measurement has been processed in the current iteration, $t_{MTT}$ is incremented by the time elapsed since the last iteration. The time step $T$ used for the EKF model discretization is computed as $T=t_{meas}-t_{MTT}^{old}$, with $t_{MTT}^{old}$ denoting the value of $t_{MTT}$ before processing the current measurement. To compensate for the sensor delay, the output of the MTT algorithm is always subject to a final EKF prediction by a time step equal to $T_{out}=t_{curr}-t_{MTT}$ with $t_{curr}$ being the current time. The EKF state is not changed during this last prediction.

\begin{figure}[t!]
    \centering
    \includegraphics[width=\columnwidth]{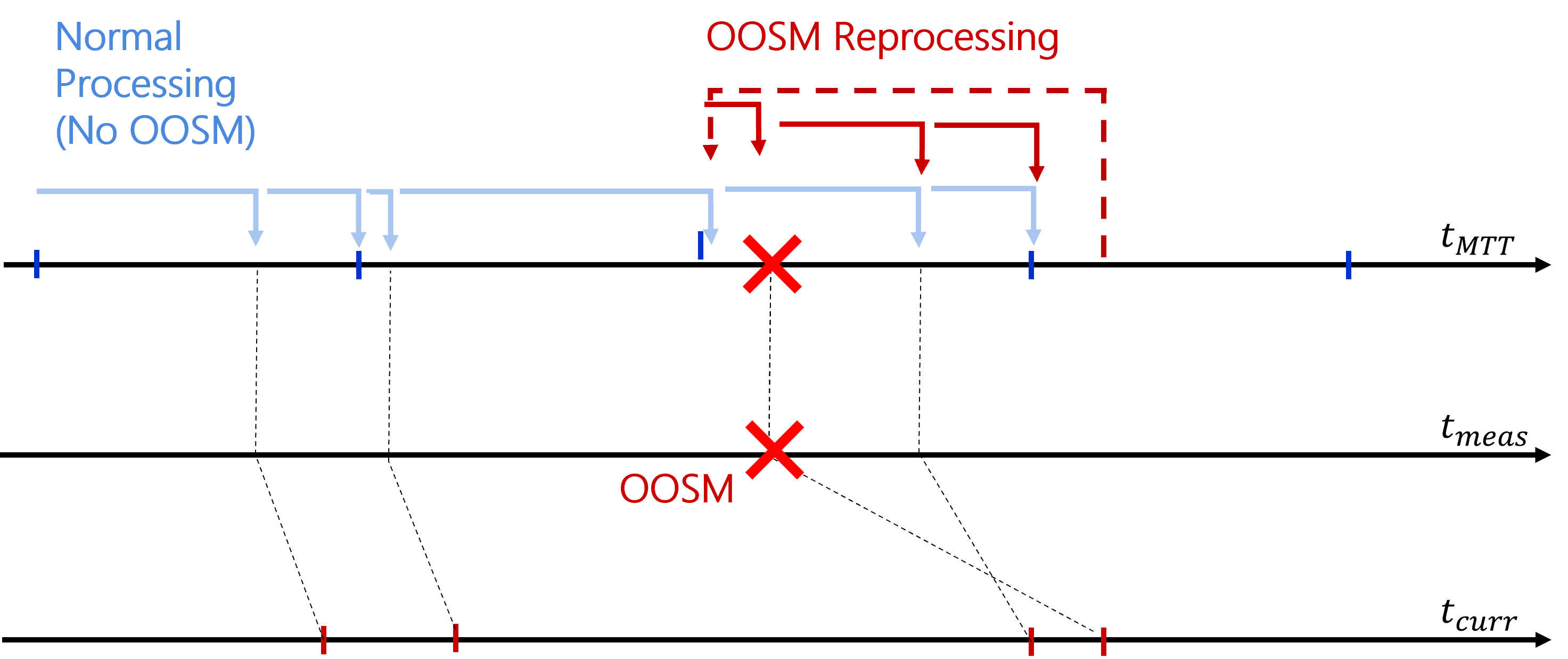}
    \caption[Scheme of the OOSM handling via Reprocessing]{Scheme of the OOSM handling via Reprocessing: while in normal operation (left) the measurements are processed in the order of arrival, when an OOSM occurs (right) the state of the MTT is reset to the last valid measurement before the OOSM timestamp and then the measurements are processed in order.}
    \label{fig:tt-oosm}
\end{figure}

If the measurement delay is large or varies greatly between the sensors, we may obtain $T<0$. This is the definition of an Out-of-Sequence Measurement (OOSM), as we need to process a measurement older than the current state estimate.

To manage OOSMs, we implemented the \textit{Reprocessing} strategy shown in Figure \ref{fig:tt-oosm}, which involves the use of an additional buffer, the \textit{Past States and Measurements Buffer} $B_{PSM}$, again implemented as an Ordered Map. After each MTT cycle update, we add a snapshot of the current Track EKF states and the corresponding input measurements (if any) to $B_{PSM}$, with $t_{meas} = t_{MTT}$ as the key. 
When an OOSM occurs, we traverse $B_{PSM}$ from the last element until we find a states and measurements pair that is older than the OOSM timestamp. In the process, we add all the encountered measurements to $B_M$. We then roll back the MTT         state, meaning the state of each track, to the snapshot from $B_{PSM}$ that precedes the OOSM, and proceed to process all the measurements added to $B_M$, including the OOSM that triggered the Reprocessing, following the standard MTT cycle.

\subsection{EKF State, Input and Measurement Function}
For each tracked target, we instantiate an EKF to estimate its dynamics by filtering the associated measurements. The EKF is composed of a state update function $\mathbf{F}$ and an output function $\mathbf{H}$ according to

\begin{equation}
        \label{eq:state_output}
        \begin{aligned}
        X_{k+1}     &= \mathbf{F}(X_k,u_k) + w_k \\
        Y_{k}       &= \mathbf{H}(X_k) + v_k
        \end{aligned}
\end{equation}
with $\mathbf{F}$ being the state transition function, $\mathbf{H}$ the measurement function, and $w_k \sim \mathcal{N}(\mathbf{0},\,\mathbf{Q}_k)$ and $v_k \sim \mathcal{N}(\mathbf{0},\,\mathbf{R}_k)$ representing the process and measurement noise model.

As the EKF state update function $\mathbf{F}$, we employ a reduced-state version of the Constant Velocity and Turn Rate (CVTR) model in its polar form \cite{roth2014ekf}. This model describes the motion of a kinematic point along a circular path, with constant linear and angular velocity, and it is widely employed in Target Tracking literature. It is described in continuous time form as
\begin{equation}
        \label{eq:cvtr_model_ct}
        \begin{aligned}
        \dot{\mathbf{X}}(t)
        &=
        \begin{bmatrix}
        \dot{x}(t) \\
        \dot{y}(t) \\
        \dot{v}(t) \\
        \dot{\theta}(t)
        \end{bmatrix}
        &=
        \begin{bmatrix}
        v(t)\cos\bigl(\theta(t)\bigr) \\
        v(t)\sin\bigl(\theta(t)\bigr) \\
        0 \\
        \omega(t) 
        \end{bmatrix}
        \end{aligned}
\end{equation}
where the state vector $X(t)=[x,y,v,\theta] \in \mathbb{R}^{4}$ describes the target pose and linear velocity in an inertial, earth-fixed reference frame. We define the turn rate $\omega$, which is the last state of the original CVTR model \cite{roth2014ekf}, as an exogenous input for our EKF, $u(t) = \hat\omega(t) \in \mathbb{R}$, hence employing a reduced-state CVTR model. 

\begin{figure}
        \centering
        \includegraphics[width=0.35\columnwidth]{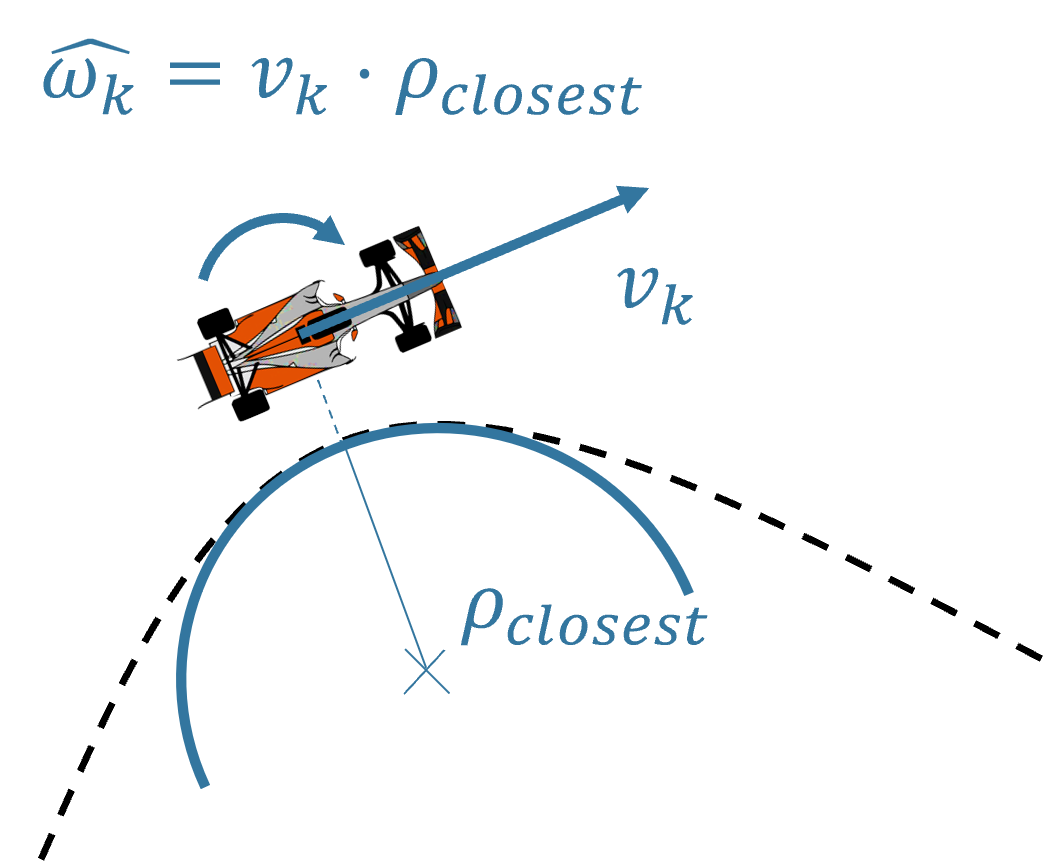}
        \caption{Computation of the Yaw Rate Exogenous Input using the vehicle estimated velocity $v_k$ and the track curvature $\rho_{closest}$}
        \label{fig:virtual_yaw_rate}
\end{figure}

The CVTR model allows for an exact discretization, which holds for any time step, enabling us to vary the time-step at runtime to accommodate non-constant sensor sampling rate and delay, as described by
\begin{equation}
        \scalebox{0.8375}{$
                \label{eq:cvtr_model_dt}
                \begin{aligned}
                \mathbf{X}(k+1) 
                &=
                \begin{bmatrix}
                \mbox{$
                    \begin{cases}
                    x_k + \frac{2v_k}{\omega_k} \sin\!\left(\frac{\omega_k T}{2}\right)
                          \cos\!\left(\theta_k+\frac{\omega_k T}{2}\right) & \omega_k \neq 0\\[6pt]
                    x_k + v_k\,T\cos\!\left(\theta_k\right)                   & \omega_k = 0
                    \end{cases}
                $}\\[8pt]
                \mbox{$
                    \begin{cases}
                    y_k - \frac{2v_k}{\omega_k} \sin\!\left(\frac{\omega_k T}{2}\right)
                          \sin\!\left(\theta_k+\frac{\omega_k T}{2}\right) & \omega_k \neq 0\\[6pt]
                    y_k + v_k\,T\sin\!\left(\theta_k\right)                   & \omega_k = 0
                    \end{cases}
                $}\\[8pt]
                v_k\\[6pt]
                \theta_k + \omega_k\,T
                \end{bmatrix}
                \end{aligned}
        $}
        \end{equation}
where the Yaw Rate Exogenous Input $\omega$ is computed by multiplying the current track velocity estimate $v_{k}$ by the curvature of a reference trajectory parallel to the track centerline. The curvature is computed from the second-order Taylor approximation of the reference trajectory at the point closest to the tracked vehicle pose, as Figure \ref{fig:virtual_yaw_rate} shows, according to:

\begin{equation}
        u_k = \hat{\omega}_k = v_k\cdot\rho_{closest}(x_k,y_k)
        \label{eq:yaw_rate_input}
\end{equation}

The rationale behind the use of this Yaw Rate Exogenous Input comes from three observations: first, a vehicle moving on a racetrack close to its performance limit, even when performing racing maneuvers like overtakes, will still have a very small incidence angle to the track centerline, as the majority of its lateral dynamics will be employed to stay within the track bounds. The second observation lies in the difficulty of estimating a state that is not measured directly but instead is the derivative of another state, again not measured directly and with a non-linear dependence on the measurements. The third observation lies in its minor role in path planning: during overtaking maneuvers the ego and opponent vehicles will be in close proximity, and therefore the opponent motion prediction will be mostly based on the opponent position, velocity and heading.

\subsection{EKF Measurement Function}

The first step of every MTT cycle is EKF Prediction, in which every track is subject to motion prediction according to \eqref{eq:cvtr_model_dt} and \eqref{eq:yaw_rate_input}. Then, we use the EKF measurement function in order to compute the predicted measurement for every track, enabling data association and EKF correction.

The EKF measurement function $\mathbf{H}(X)$ is time-varying and size-varying, as the shape of the measurement depends on the type of sensor employed: it can be $Y_L=[x,y] \in \mathbb{R}^2$ or $Y_R=[x,y,R_r] \in \mathbb{R}^3$. For both LiDAR and RADAR measurements, we apply the transformation from local (ego) to global (earth-fixed) reference frame for the ${x,y}$ position measurements, in order to have them in the same reference frame as the EKF model.

The LiDAR measurement function is linear, as after the reference frame transformation the LiDAR directly measures the first two states, giving $Y_{k}=H_L\cdot X_{k}$, from which derives:
\begin{equation}
        H_L = 
        \begin{bmatrix}
        1 & 0 & 0 & 0 \\
        0 & 1 & 0 & 0
        \end{bmatrix}
        \label{eq:H_Lidar}
\end{equation}

\begin{figure}
        \centering
        \includegraphics[width=0.55\columnwidth]{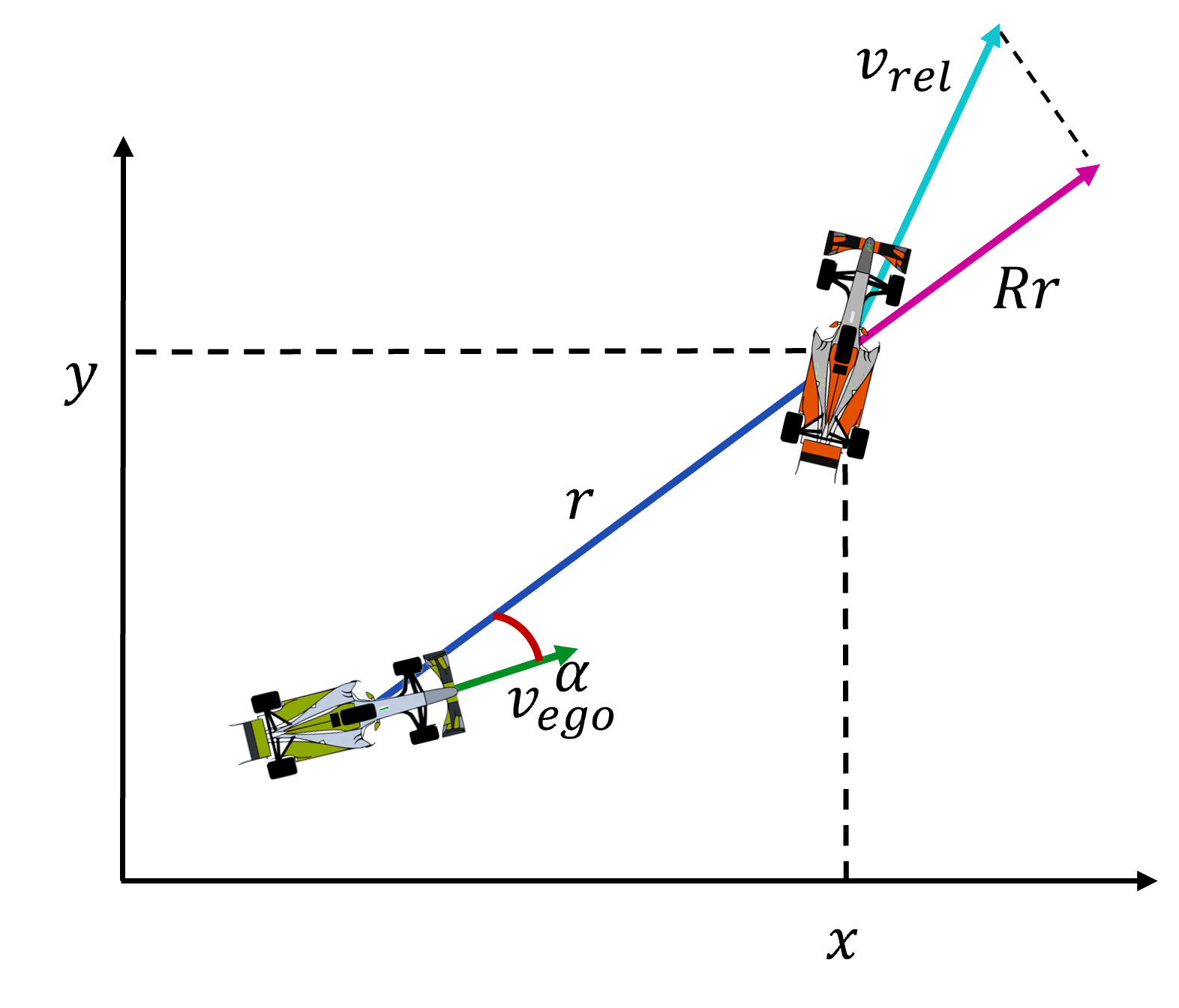}
        \caption[Scheme of the range-rate measurement]{Representation of the components of the RADAR Doppler measurement function: the range-rate is the projection of the relative velocity $v_{rel}$ over the Line-of-Sight angle $\alpha$.}        
        \label{fig:range_Rate}
\end{figure}

The RADAR measurement function instead is non-linear and time-varying as the range-rate represents the relative ego-target velocity expressed in the ego reference frame and projected on the Line-of-Sight defined by the bearing angle $\alpha$, as Figure \ref{fig:range_Rate} shows. Therefore, the RADAR measurement function depends not only on the target states but also on the ego velocity and heading, which are assumed to be known. Given the knowledge of the ego velocity and heading ${v_{e}},\theta_{e}$, in the hypothesis of small ego sideslip angle, the range-rate equation is
\begin{multline}
        R_r = [-v\cos(\theta)\sin(\theta_{e}) + v\sin(\theta)\cos(\theta_{e})]\sin(\alpha)\\
         + [v\cos(\theta)\cos(\theta_{e}) + v\sin(\theta)\sin(\theta_{e})-v_{e}]\cos(\alpha)
         \label{eq:range_rate}
\end{multline}
and its corresponding linearized measurement matrix can be derived as:
\begin{equation}
        H_R = 
        \begin{bmatrix}
        1 & 0 & 0               & 0 \\
        0 & 1 & 0               & 0 \\
        0 & 0 & \frac{dR_r}{dv} & \frac{dR_r}{d\theta}
        \end{bmatrix}
        \label{eq:H_RADAR}
\end{equation}

\subsection{Data Association and Track Management}

During the MTT prediction step, we apply \eqref{eq:state_output} to compute both the predicted state $\hat{X}_{k+1}$ and its corresponding predicted output $\hat{Y}$ for every track, using \eqref{eq:H_Lidar} or \eqref{eq:H_RADAR} depending on the sensor type. We also update the current state and output error covariance estimation according to:

\begin{equation}
        \begin{aligned}
        \hat{X}_{k}    &= \mathbf{F}(X_{k-1},u_k)      \\
        \hat{Y}_{k}    & =\mathbf{H}(\hat{X}_k)                \\
        F_{k}          &= \left .\frac{d\mathbf{F}}{dx}\right |_{\hat{X}_k}       \\
        P_{k}          &= F_k\cdot P_{k-1}\cdot F_k^T + Q                 \\
        S_{k}          &= H_{k}\cdot P_k\cdot H_{k}^T + R         
        \end{aligned}
        \label{eq:ekf_prediction}
\end{equation}
where $F$ represents the state linearization around the current prediction, $P$ the state error covariance estimate and $S$ the output error covariance estimate, while $Q$ and $R$ represent the model and output noise covariance matrices.

We associate the incoming measurements to the predicted track position by computing the Mahalanobis Distance for every track-measurement pair, defined as $D_m$. To ensure a uniform distance computation across different sensor types, we only employ the $\{x,y\}$ position of the RADAR measurement and the relative submatrix of $S$ according to:
\begin{equation}
        \begin{aligned}
        E_m       &= \begin{bmatrix}
                        \hat{x}-x_{meas},\quad \hat{y}-y_{meas}
                \end{bmatrix} \\
        D_m       &= E_m \cdot (S_{1:2,\,1:2})^{-1} \cdot E_m^T
        \end{aligned}
        \label{eq:Mahalanobis}
\end{equation}

We evaluate the Mahalanobis Distance for every track-measurement pair, building a cost matrix. We impose an upper gating threshold on the computed Mahalanobis distance in order to inhibit the association of measurements that are too far away from the corresponding track, and then we use the Munkres Algorithm to find the track-measurement assignment with the lowest overall cost. The use of the Mahalanobis distance for data association allows the data association process to encompass information about the current track state estimation confidence: a track that has not been corrected for long will have a larger association region to reflect the larger uncertainty associated with the EKF position estimate.

After the Data Association step, measurements associated with a track will be used to correct the EKF state prediction, while every measurement that has not been associated with an existing track will be used to spawn a new track.

\begin{figure}
        \centering
        \includegraphics[width=.9\columnwidth]{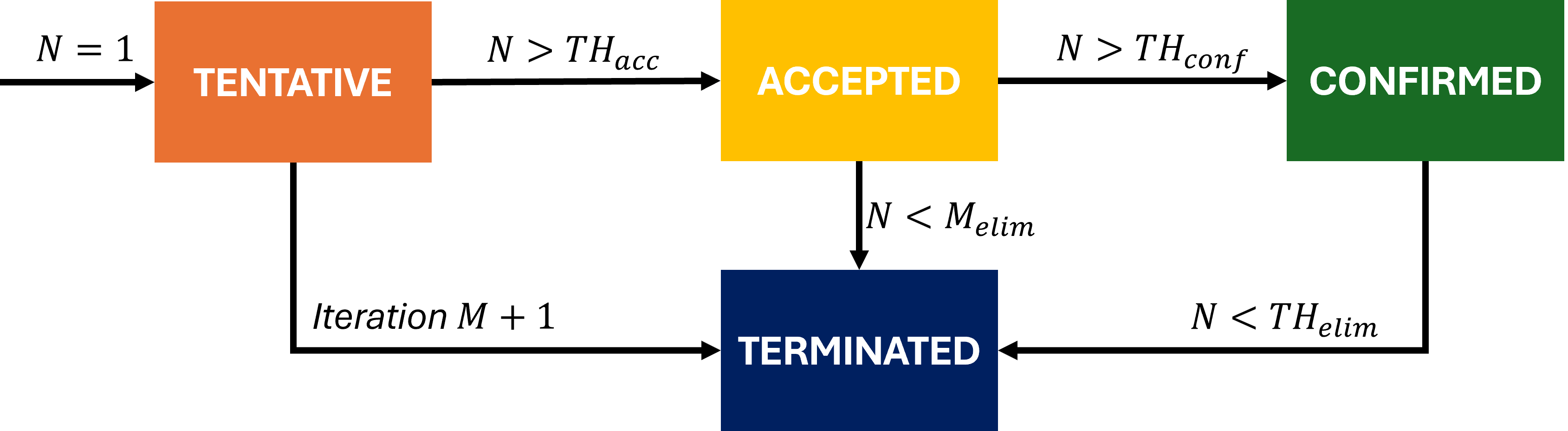}
        \caption{Track Management Finite State Machine states and transitions}
        \label{fig:FSM}
\end{figure}

For every new track, we instantiate a Finite-State Machine (FSM), shown in Figure \ref{fig:FSM}, with four states: \textit{Tentative, Accepted, Confirmed} and \textit{Terminated}. Every track FSM is initialized in the \textit{Tentative} state, with the \textit{Terminated} state being the terminal state. Only \textit{Confirmed} tracks represent the output of the Target Tracking algorithm, while \textit{Accepted} tracks represent an intermediate state.

The state transition function of the FSM depends on the so-called \textit{M/N Logic}, meaning that state transitions are enabled if the corresponding track has received at least \textit{N} measurements in the last \textit{M} MTT cycles. We define two upper thresholds, $TH_{acc}$ and $TH_{conf}>TH_{acc}$ for the transition from \textit{Tentative} to \textit{Accepted} and from \textit{Accepted} to \textit{Confirmed} respectively. A lower threshold, $TH_{elim}<TH_{acc}$, is used to transition tracks from either the \textit{Accepted} or \textit{Confirmed} states to the \textit{Terminated} state if not met, effectively eliminating them. The elimination of \textit{Tentative} tracks happens if, once \textit{M} measurements have been received, the track has not been \textit{Accepted}.

\subsection{EKF Correction and Initialization}
The last step of the MTT cycle is the actual EKF correction, in which we use the measurement associated with every track (except \textit{Terminated} tracks) to correct the estimation of its dynamical state using the EKF update equations as in
\begin{equation}
        \begin{aligned}
        E_{k} &= (Y_k - \hat{Y}_k) \\
        L_{k} &= P_k\cdot H_{k}^T \cdot S^{-1}_k \\
        X_{k} &= X_{k-1} + L_k \cdot E_k
        \end{aligned}  
        \label{eq:ekf_correction}
\end{equation}
where $E$ is the measurement error or innovation of the associated measurement $Y$ and $L$ is the Kalman Gain.

This EKF correction is not applied to the first associated measurement of a track, which determines the track creation and is initialized with $v=0,\theta=0$, nor for the second associated measurement: to speed up the EKF convergence, we perform the two-step initialization procedure described in \cite{bar2004estimation}.
During this procedure we compute the initialization velocity and heading as:
\begin{equation}
        \begin{aligned}
                v      &= \sqrt{(x_2-x_1)^2+(y_2-y_1)^2} \\
                \theta &= \atantwo{(y_2-y_1,x_2-x_1)}
        \end{aligned}
        \label{eq:ekf_two_step_init}
\end{equation}

This procedure quickly initializes the track's velocity and heading to values that, although noisy, enable faster convergence to the true values compared to the zero initialization.

\section{EXPERIMENTAL RESULTS}\label{sec:experimental_results}

\begin{figure}[b]
        \centering
        \includegraphics[width=.7\columnwidth]{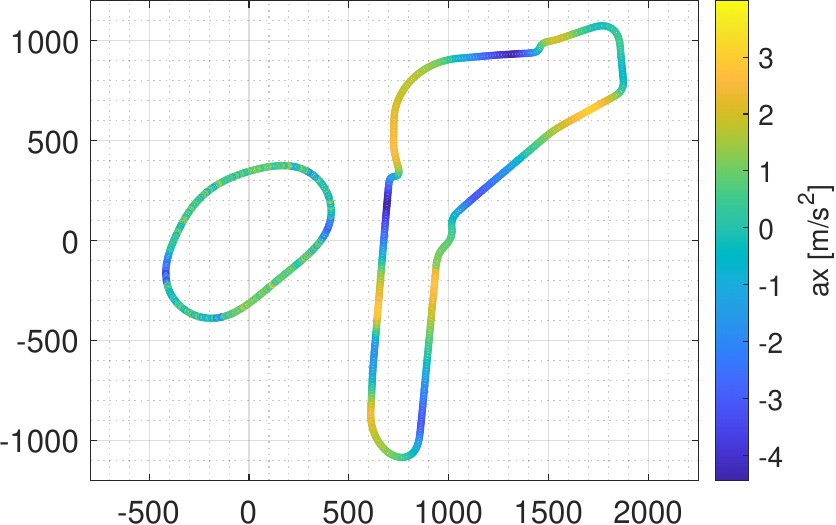}
        \caption{Map and acceleration profile of the two racing circuits used for this work: on the left, the Las Vegas Motor Speedway, on the right the Autodromo Nazionale Monza. Distances are expressed in meters.}
        \label{fig:acceleration_map}
\end{figure}

The vehicle used for the experimental data collection and algorithm validation is the Dallara AV-21 shown in Figure \ref{fig:intro}. It is a single-seater autonomous race car equipped with three Luminar Hydra LiDAR sensors, positioned to provide almost 360° coverage around the vehicle, and a single forward-facing APTIV ESR 2.5 RADAR sensor. The vehicle is equipped with two NovAtel PwrPak7D RTK INS/GNSS receivers which act both as the vehicle localization source for the algorithm operations as well as the opponent ground truth source, as all IAC cars follow the same specification.

We evaluate the algorithm's performance on two datasets: the first is an experimental dataset taken at the Las Vegas Motor Speedway during the 2023 IAC event, consisting of 15 overtaking maneuvers captured from both the overtaking and overtaken vehicles' perspectives, up to 160 to 275 km/h. Due to the oval racetrack, as well as the regulationsimposing specific speed brackets for the overtaken vehicle, this dataset is characterized by long sequences at almost constant speeds. We generated a second dataset using the experimental data acquired during the PoliMOVE qualifying lap at the 2023 IAC event at the Autodromo Nazionale Monza, by simulating a chasing vehicle following a leader using the recorded trajectories of the real vehicle. We added noise to the ideal sensor measurements with a zero-mean Gaussian model, following the noise characterization of the real sensor data with respect to the ground truth. Figure \ref{fig:acceleration_map} shows the different acceleration profile of the two datasets, showing the highly variable vehicle speed on the Monza track.

The online ROS2 implementation yields an average computational time under 1 ms, which is negligible with respect to the LiDAR processing latency of 26ms \cite{cellina2025lidar} as well as the estimated RADAR latency of 100ms. The latency compensation via the final EKF state prediction compensates the end-to-end latency of the VDT pipeline.

\begin{figure}[t]
    \centering
    \includegraphics[width=.95\columnwidth]{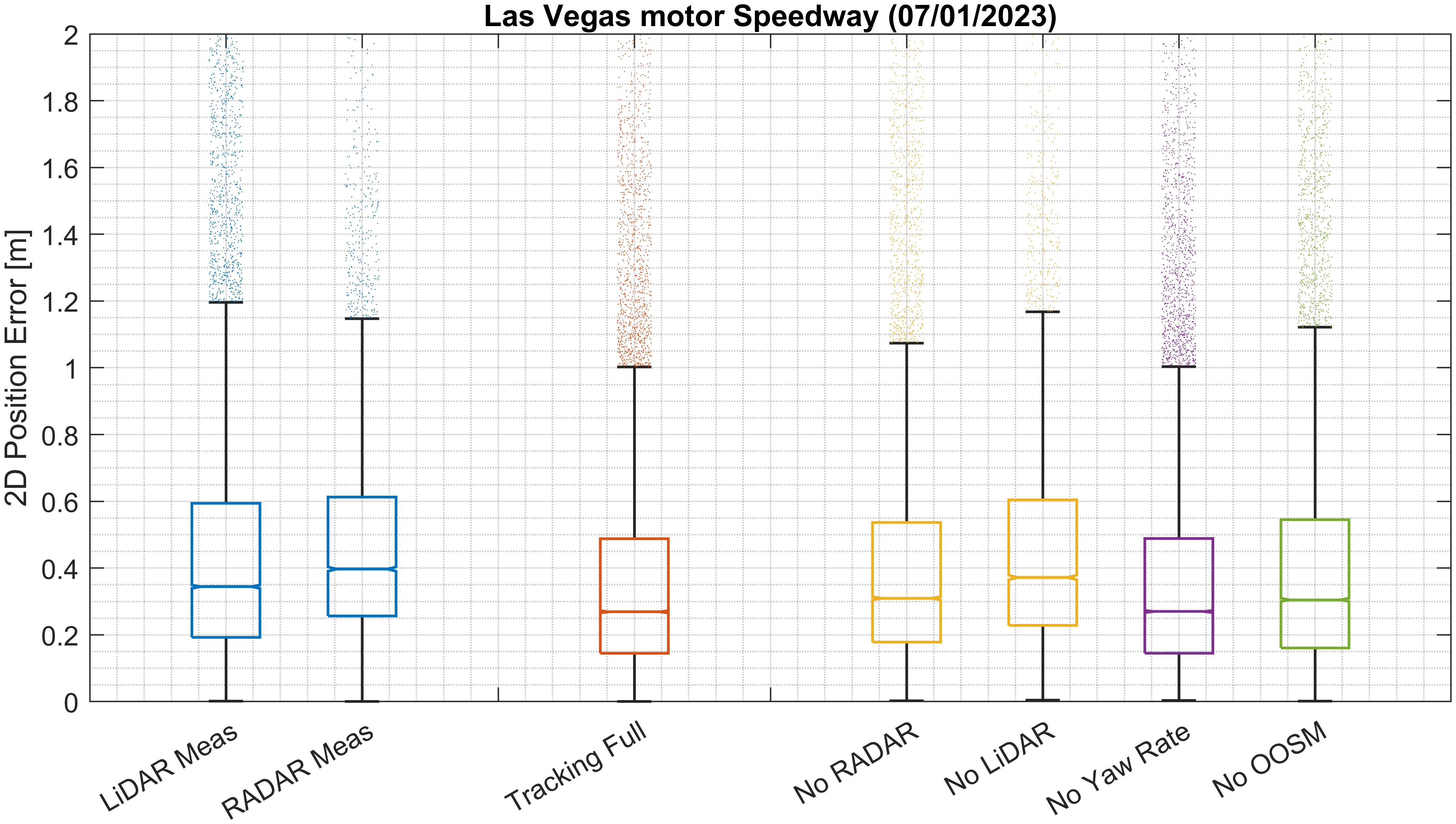}
    \caption{2D position error ablation study on the LVMS dataset}
    \label{fig:res-lvms}
\end{figure}

\begin{figure}[b]
    \centering
    \includegraphics[width=.95\columnwidth]{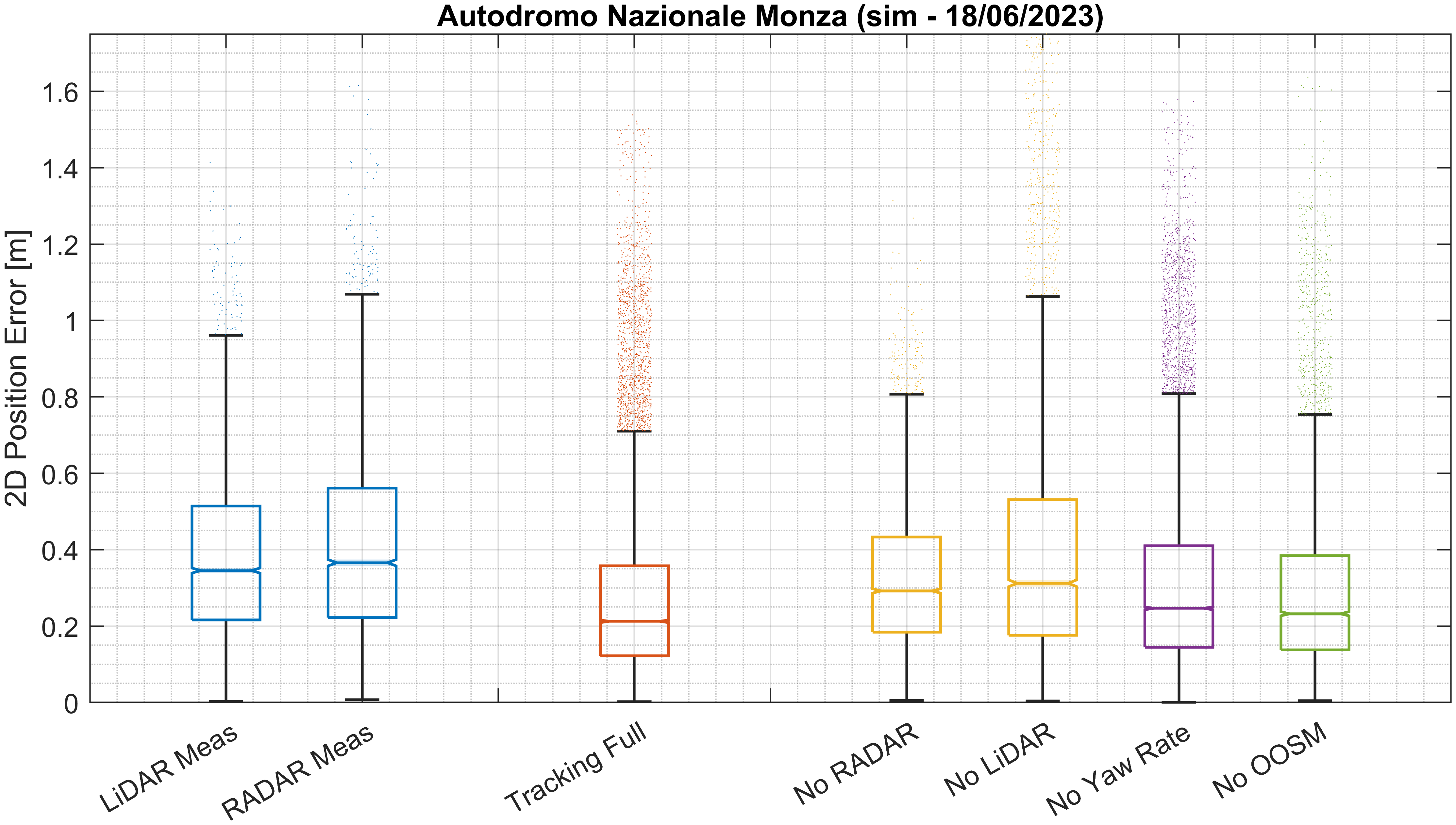}
    \caption{2D position error ablation study on the Simulated Monza dataset}
    \label{fig:res-monza}
\end{figure}

We compute the 2D position error to evaluate the performance of the tracking algorithm, and then we perform an ablation study by disabling each component one at a time in order to understand their contribution to the overall performance. Figure \ref{fig:res-lvms} displays the distribution of the raw measurements and of the Tracking state estimation 2D position error computed with respect to the GNSS ground truth for the LVMS dataset. The figure shows how the LiDAR has an overall lower 2D position estimation error compared with RADAR, although with higher variance, whereas sensor-fusion tracking attains a lower average error, evidencing the filter's noise-reduction effect. The deactivation of each sensor increases the median, with the RADAR-only tracking being noisier than the LiDAR-only tracking, reflecting the larger position uncertainty of the latter. The deactivation of the Yaw Rate Exogenous Input, meaning the estimation of $\omega$ as the fifth EKF state using the full-state CVTR model, does not significantly reduce the filter's performance, while the disabling of OOSM management, by simply discarding OOSMs when they are present, greatly degrades tracking accuracy, as half of the measurements are discarded.

Figure \ref{fig:res-monza} shows the result of the 2D position error ablation study for the simulated Monza dataset. While the full tracking error precision, as well as the trend in the ablation study, is comparable to the experimental data, the contribution of the Virtual Yaw Rate Exogenous Input to the reduction of the state estimation error is greater in the Monza dataset than it was in the LVMS dataset. This can be explained by considering the higher vehicle yaw rate dynamics in a road course circuit presenting more abrupt direction changes compared with an oval raceway. In general, the positive effect of direct state measurement or input (velocity via RADAR Doppler range-rate measurements, and yaw rate via the Exogenous Input) is more distinguishable in the simulated Monza dataset, due to the more aggressive acceleration profile.

\section{CONCLUSIONS}\label{sec:conclusions}

In this paper, we present a latency-aware, EKF-Based Multi-Target Tracking algorithm that integrates LiDAR and RADAR measurements, explicitly incorporating track curvature in the prediction step and RADAR range-rate information in the measurement model. By managing Out-Of-Sequence Measurements through state and measurement reprocessing, the proposed method can effectively handle the different delays arising from multi-modal fusion. Experimental validation on real-world data from the Las Vegas Motor Speedway, as well as the simulated Monza dataset, demonstrates robust and reliable estimation performance under widely varying conditions and acceleration profiles. Ablation studies on both datasets show the contribution of each module to reducing the estimation error.

Future work will extend this approach to include camera-based perception and V2V communication, while also investigating advanced tracking strategies such as Interactive Multiple Models (IMM) to further enhance performance under complex racing maneuvers like heavy braking. 




\bibliographystyle{IEEEtran}
\bibliography{bibliography/IEEEabrv,bibliography/PAPER_2025_IAC_LIDAR_RADAR_TRACKING}

\end{document}